\ifdefined\pdfminorversion\pdfminorversion=4\fi
\documentclass{static/ifacconf}

\usepackage{graphicx}      %
\usepackage{natbib}        %
\usepackage{amsmath,amssymb}
\usepackage{xcolor}
\usepackage{import}        %
\usepackage{bm,mathtools,xspace}
\usepackage[nolist]{acronym}  %
\usepackage{comment}
\usepackage{eso-pic}       %

\newif\ifdraftnotes
\draftnotesfalse   %
\newcommand{\redcomment}[1]{\ifdraftnotes\textcolor{red}{#1}\fi}   %
\newcommand{\lvwidth}[1]{}

\ifdraftnotes

\else
  \excludecomment{draftitemize}
  \excludecomment{draftenumerate}
\fi

\begin{document}
\AddToShipoutPictureFG*{\AtPageLowerLeft{\put(\LenToUnit{\dimexpr1in+\oddsidemargin\relax},\LenToUnit{\dimexpr7mm+\voffset\relax}){\sbox0{\fbox{\parbox{\dimexpr\textwidth-2\fboxsep-2\fboxrule\relax}{\footnotesize This work has been submitted to IFAC for possible publication.}}}\makebox[0pt][l]{\raisebox{\dp0}{\usebox0}}}}}

\begin{acronym}
	\acro{rnn}[RNN]{Recurrent Neural Network}
	\acroplural{rnn}[RNN]{Recurrent Neural Networks}
	\acro{narx}[NARX-NN]{Nonlinear Autoregressive Exogenous Neural Network}
	\acroplural{narx}[NARX-NN]{Nonlinear Autoregressive Exogenous Neural Networks}
	\acro{gru}[GRU]{{Gated Recurrent Unit}}
	\acroplural{gru}[GRU]{{Gated Recurrent Units}}
	\acro{lstm}[LSTM]{Long Short-Term Memory}
	\acro{ann}[NN]{Neural Network}	
	\acroplural{ann}[NN]{Neural Networks}
	\acro{ffnn}[FFNN]{Feedforward Neural Network}
	\acroplural{ffnn}[FFNNs]{Feedforward Neural Networks}
	\acro{pinn}[PINN]{Physics-Informed Neural Network}
	\acroplural{pinn}[PINN]{Physics-Informed Neural Networks}
	\acro{gp}[GP]{Gaussian Process}
    \acro{lgp}[LGP]{Lagrangian Gaussian Process}
    \acro{hgp}[HGP]{Hamiltonian Gaussian Process}
	\acroplural{gp}[GPs]{Gaussian Processes}
    \acroplural{lgp}[LGPs]{Lagrangian Gaussian Processes}
	\acro{knn}[$K$NN]{$K$-Nearest-Neighbors}
	\acro{ilc}[ILC]{Iterative Learning Control}
	\acro{ili}[ILI]{Iterative Learning Identification}
	\acro{rls}[RLS]{Recursive Least Squares}
	\acro{rc}[RC]{Repetitive Control}
	\acro{rl}[RL]{{Reinforcement Learning}}
	\acro{daoc}[DAOC]{{Direct Adaptive Optimal Control}}
	\acro{ml}[ML]{maschinelles Lernen}
	\acro{lwpr}[LWPR]{{Locally Weighted Projection Regression}}
	\acro{svm}[SVM]{{Support Vector Machine}}
	\acro{mcmc}[MCMC]{{Markov chain Monte Carlo}}
	\acro{ad}[AD]{{Automatic Differentiation}}
	\acro{gmm}[GMM]{{Gaussian Mixture Models}}
	\acro{rkhs}[RKHS]{{Reproducing Kernel Hilbert Spaces}}
	\acro{rbf}[RBF]{{Radial Basis Function}}
	\acro{rbfnn}[RBF-NN]{{Radial Basis Function Neural Network}}
	\acroplural{rbfnn}[RBF-NN]{{Radial Basis Function Neural Networks}}
	\acro{node}[{Neural} ODE]{{Neural Ordinary Differential Equation}}
	\acroplural{node}[{Neural} ODEs]{{Neural Ordinary Differential Equations}}
	\acro{pdf}[PDF]{Probability Density Function}
	\acro{pca}[PCA]{Principal Component Analysis}
	\acro{lnn}[LNN]{Lagrangian Neural Network}
	\acroplural{lnn}[LNNs]{Lagrangian Neural Networks}
	\acro{hnn}[HNN]{Hamiltonian Neural Network}
	\acroplural{hnn}[HNNs]{Hamiltonian Neural Networks}

	\acro{kf}[KF]{Kalman Filter}	
	\acro{ekf}[EKF]{Extended Kalman Filter}
	\acro{nekf}[NEKF]{Neural Extended Kalman Filter}
	\acro{ukf}[UKF]{Unscented Kalman Filter}
	\acro{pf}[PF]{Particle Filter}
	\acro{mhe}[MHE]{{Moving Horizon Estimation}}
	\acro{rpe}[RPE]{{Recursive Predictive Error}}
	\acro{slam}[SLAM]{{Simultaneous Location and Mapping}}
	
	\acro{mftm}[MFTM]{Magic Formula Tire Model}
	\acro{mbs}[MBS]{Multi-Body Simulation}
	\acro{lti}[LTI]{Linear Time-Invariant}
	\acro{cog}[COG]{Center Of Gravity}
	\acro{ltv}[LTV]{Linear Time-Variant}
	\acro{siso}[SISO]{Single Input Single Output}
	\acro{mimo}[MIMO]{Multiple Input Multiple Output}
	\acro{psd}[PSD]{Power Spectral Density}
	\acroplural{psd}[PSD]{Power Spectral Densities}
	\acro{cf}[CF]{Coordinate Frame}
	\acroplural{cf}[CF]{Coordinate Frames}
	\acro{phs}[PHS]{Port-Hamiltonian System}

	\acro{pso}[PSO]{Particle Swarm Optimization}
	\acro{sqp}[SQP]{Sequentielle Quadratische Programmierung}
	\acro{svd}[SVD]{Singular Value Decomposition}
	\acro{ode}[ODE]{Ordinary Differential Equation}
	\acroplural{ode}[ODEs]{Ordinary Differential Equations}
	\acro{pde}[PDE]{Partial Differential Equation}
	\acro{sde}[SDE]{stochastic differential equation}
	\acroplural{sde}[SDEs]{Stochastic Differential Equations}

	\acro{nmse}[NMSE]{Normalized Mean Squared Error}
	\acro{mse}[MSE]{Mean Squared Error}
	\acro{rmse}[RMSE]{Root Mean Squared Error}
	\acro{wrmse}[wRMSE]{weighted \ac{rmse}}
	
	\acro{mpc}[MPC]{{Model Predictive Control}}
	\acro{nmpc}[NMPC]{Nonlinear Model Predictive Control}
	\acro{lmpc}[LMPC]{Learning Model Predictive Control}
	\acro{ltvmpc}[LTV-MPC]{\acl{ltv} Model Predictive Control}
	\acro{ndi}[NDI]{Nonlinear Dynamic Inversion}
	\acro{ac}[AC]{Adhesion Control}
	\acro{esc}[ESC]{Electronic Stability Control}
	\acro{ass}[ASS]{Active Suspension System}
	\acro{trc}[TRC]{Traction Control}
	\acro{abs}[ABS]{Anti-Lock Brake System}
	\acro{gcc}[GCC]{Global Chassis Control}
	\acro{ebs}[EBS]{Electronic Braking System}
	\acro{adas}[ADAS]{Advanced Driver Assistance Systems}
	\acro{cm}[CM]{{Condition Monitoring}}
	\acro{hil}[HiL]{{Hardware-in-the-Loop}}
	\acro{siso}[SISO]{Single Input Single Output}
	\acro{mimo}[MIMO]{Multiple Input Multiple Output}

	\acro{irw}[IRW]{Independently Rotating Wheels}
	\acro{dirw}[DIRW]{Driven \acl{irw}}
	\acro{imes}[imes]{{Institute of Mechatronic Systems}}
	\acro{db}[DB]{Deutsche Bahn}
	\acro{ice}[ICE]{Intercity-Express}
	
	\acro{fmi}[FMI]{{Functional Mock-up Interface}}
	\acro{fmu}[FMU]{{Functional Mock-up Unit}}
	\acro{doi}[DOI]{{Digital Object Identifier}}

	\acro{ra}[RA]{Research Area}
	\acroplural{ra}[RA]{Research Areas}
	\acro{wp}[WP]{Work Package}
	\acroplural{wp}[WP]{Work Packages}
	
	\acro{fb}[FB]{Forschungsbereich}
	\acroplural{fb}[FB]{Forschungsbereiche}
	\acro{ap}[AP]{Arbeitspaket}
	\acroplural{ap}[AP]{Arbeitspakete}
	\acro{abb}[Abb.]{Abbildung}
	\acro{luis}[LUIS]{Leibniz Universit"at IT Services}
	
	\acro{res}[RES]{Renewable Energy Sources}
	\acro{pkw}[PKW]{Personenkraftwagen}
	\acro{twipr}[TWIPR]{{Three-Wheeled Inverted Pendulum Robot}}
	\acro{tlr}[TLR]{Two-Link Robot}
	
	\acro{pmcmc}[PMCMC]{particle Markov chain Monte Carlo}
	\acro{rbpf}[RBPF]{Rao-Blackwellized Particle Filter}
	\acro{pf}[PF]{Particle Filter}
	\acro{ps}[PS]{Particle Smoother}
	\acro{smc}[SMC]{Sequential Monte Carlo}
	\acro{csmc}[cSMC]{conditional SMC}
	\acro{mh}[MH]{Metropolis Hastings}
	\acro{em}[EM]{Expectation Maximization}
	\acro{slam}[SLAM]{Simultaneous Location and Mapping}
	\acro{dof}[DOF]{Degree of Freedom}
	\acroplural{dof}[DOF]{Degrees of Freedom}
	\acro{pg}[PG]{Particle Gibbs}
	\acro{pgas}[PGAS]{Particle Gibbs with Ancestor Sampling}
	\acro{mpgas}[mPGAS]{marginalized Particle Gibbs with Ancestor Sampling}
	\acro{hmm}[HMM]{hidden Markov model}
	\acro{eom}[EOM]{Equations of Motion}
	\acro{phs}[PHS]{Port-Hamiltonian System}
	\acro{pmmh}[PMMH]{particle marginal Metropolis--Hastings}
	
	\acro{sbi}[SBI]{simulation-based inference}
	\acro{sda}[SDA]{score-based data assimilation}
	\acro{npe}[NPE]{Neural Posterior Estimation}
	\acro{nle}[NLE]{Neural Likelihood Estimation}
	\acro{nre}[NRE]{Neural Ratio Estimation}
	\acro{abc}[ABC]{Approximate Bayesian Computation}
	\acro{npse}[NPSE]{Neural Posterior Score Estimation}
	
	\acro{vi}[VI]{Variational Inference}
	\acro{kl}[KL]{Kullberg-Leibler}
	\acro{is}[IS]{Importance Sampling}
	\acro{dps}[DPS]{Diffusion Posterior Sampling}

	\acro{emps}[EMPS]{Electro-Mechanical Positioning System}
	
	\acro{ilc}[ILC]{Iterative Learning Control}
	\acro{ddilc}[DD-ILC]{Data-Driven Iterative Learning Control}
	\acro{dilc}[DILC]{Dual Iterative Learning Control}
	\acro{iml}[IML]{Iterative Model Learning}
	\acro{noilc}[NO-ILC]{Norm-Optimal Iterative Learning Control}
	\acro{gilc}[G-ILC]{Gradient Iterative Learning Control}

	\acro{bilbo}[BILBO]{Balancing Intelligent Learning roBOt}
	
	\acro{llm}[LLM]{Large Language Model}
    \acroplural{llm}[LLMs]{Large Language Models}

    \acro{rope}[RoPE]{rotary position embedding}
	\acro{gfe}[GFE]{Gaussian Fourier embedding}
	
\end{acronym}

\newcommand{\methodname}{JEDI}
\newcommand{\methodnamelong}{Joint state--parameter Estimation via Diffusion}

\newcommand{\qvec}[1]{\mathbf{#1}}
\newcommand{\qmat}[1]{\mathbf{#1}}
\newcommand{\idx}[1]{_{\mathrm{#1}}}

\newcommand{\qRealNumbers}{\mathbb{R}}
\newcommand{\qPositiveRealNumbers}{\mathbb{R}_{\geq 0}}
\newcommand{\qNaturalNumbersZero}{\mathbb{N}_{\geq 0}}
\newcommand{\qNaturalNumbersPos}{\mathbb{N}_{>0}}
\newcommand{\qNaturalNumbers}{\mathbb{N}}
\newcommand{\foralljinN}{\forall j \in \qNaturalNumbersZero, \quad}
\newcommand{\foralljgeqN}{\forall j \in \qNaturalNumbersPos, \quad}
\newcommand{\qBLT}{\mathcal{T}^{\mathrm{BLT}}}
\newcommand{\qNormal}{\mathcal{N}}

\newcommand\invChi{\mathop{\mbox{Scale-inv-$\chi^2$}}}

\newcommand{\qa}{\qvec{a}}
\newcommand{\qb}{\qvec{b}}
\newcommand{\qd}{\qvec{d}}
\newcommand{\qe}{\qvec{e}}
\newcommand{\qf}{\qvec{f}}
\newcommand{\qg}{\qvec{g}}
\newcommand{\qh}{\qvec{h}}
\newcommand{\qi}{\qvec{i}}
\newcommand{\qj}{\qvec{j}}
\newcommand{\qk}{\qvec{k}}
\newcommand{\ql}{\qvec{l}}
\newcommand{\qm}{\qvec{m}}
\newcommand{\qn}{\qvec{n}}
\newcommand{\qo}{\qvec{o}}
\newcommand{\qp}{\qvec{p}}
\newcommand{\qr}{\qvec{r}}
\newcommand{\qs}{\qvec{s}}
\newcommand{\qt}{\qvec{t}}
\newcommand{\qu}{\qvec{u}}
\newcommand{\qv}{\qvec{v}}
\newcommand{\qw}{\qvec{w}}
\newcommand{\qx}{\qvec{x}}
\newcommand{\qy}{\qvec{y}}
\newcommand{\qz}{\qvec{z}}
\newcommand{\qem}{\qvec{e}^{\mathrm{m}}}

\newcommand{\qA}{\qvec{A}}
\newcommand{\qB}{\qvec{B}}
\newcommand{\qC}{\qvec{C}}
\newcommand{\qD}{\qvec{D}}
\newcommand{\qE}{\qvec{E}}
\newcommand{\qF}{\qvec{F}}
\newcommand{\qG}{\qvec{G}}
\newcommand{\qH}{\qvec{H}}
\newcommand{\qI}{\qvec{I}}
\newcommand{\qJ}{\qvec{J}}
\newcommand{\qK}{\qvec{K}}
\newcommand{\qL}{\qvec{L}}
\newcommand{\qM}{\qvec{M}}
\newcommand{\qN}{\qvec{N}}
\newcommand{\qO}{\qvec{O}}
\newcommand{\qP}{\qvec{P}}
\newcommand{\qQ}{\qvec{Q}}
\newcommand{\qR}{\qvec{R}}
\newcommand{\qS}{\qvec{S}}
\newcommand{\qT}{\qvec{T}}
\newcommand{\qU}{\qvec{U}}
\newcommand{\qV}{\qvec{V}}
\newcommand{\qW}{\qvec{W}}
\newcommand{\qX}{\qvec{X}}
\newcommand{\qY}{\qvec{Y}}
\newcommand{\qZ}{\qvec{Z}}

\newcommand{\quv}{\bar{\qu}}
\newcommand{\qyv}{\bar{\qy}}
\newcommand{\qxv}{\bar{\qx}}
\newcommand{\qZero}{\qvec{0}}
\newcommand{\qdu}{\boldsymbol{\Delta}\qu}
\newcommand{\qdy}{\boldsymbol{\Delta}\qy}
\newcommand{\qep}{\hat{\qe}}
\newcommand{\qyp}{\hat{\qy}}

\newcommand{\qILCDesign}{\boldsymbol{D}}
\newcommand{\qIMLDesign}{\boldsymbol{\hat{D}}}
\newcommand{\qLIML}{\hat{\qL}}
\newcommand{\qeR}{\qe_\mathrm{R}}

\newcommand{\qff}{\boldsymbol{f}}
\newcommand{\qpf}{\boldsymbol{p}}
\newcommand{\qmf}{\boldsymbol{m}}
\newcommand{\qXf}{\boldsymbol{A}}

\newcommand{\qLift}{\mathcal{L}}
\newcommand{\qLu}{\boldsymbol{\mathcal{L}\idx{u}}}
\newcommand{\qLm}{\boldsymbol{\mathcal{L}\idx{m}}}
\newcommand{\qUb}{\bar{\qU}}
\newcommand{\qyh}{\hat{\qy}}
\newcommand{\qeh}{\hat{\qe}}
\newcommand{\qLh}{\hat{\qL}}
\newcommand{\qDeh}{\hat{\boldsymbol{\mathcal{D}}}}
\newcommand{\qDe}{{\boldsymbol{\mathcal{D}}}}
\newcommand{\qMt}{\tilde{\qM}}
\newcommand{\qPt}{\tilde{\qP}}
\newcommand{\qLt}{\tilde{\qL}}
\newcommand{\qQt}{\tilde{\qQ}}
\newcommand{\qSt}{\tilde{\qS}}
\newcommand{\qWt}{\tilde{\qW}}
\newcommand{\qRt}{\tilde{\qR}}
\newcommand{\qut}{\tilde{\qu}}
\newcommand{\qyt}{\tilde{\qy}}
\newcommand{\qet}{\tilde{\qe}}
\newcommand{\qzt}{\tilde{\qz}}
\newcommand{\qrt}{\tilde{\qr}}
\newcommand{\qXt}{\tilde{\qX}}

\newcommand{\qTb}{\bar{\qT}}
\newcommand{\qVb}{\bar{\qV}}

\newcommand{\qQh}{\hat{\qQ}}
\newcommand{\qSh}{\hat{\qS}}
\newcommand{\qXh}{\hat{\qX}}
\newcommand{\qWh}{\hat{\qW}}

\newcommand{\Lfunc}{\mathscr{L}}       %
\newcommand{\Kfunc}{\mathscr{K}}
\newcommand{\KLfunc}{\Kfunc\negthinspace\negthinspace\Lfunc}

\newcommand{\qTwoNorm}[1]{\left\| {#1} \right\|_{2}}
\newcommand{\qInfNorm}[1]{\left\| {#1} \right\|_\infty}
\newcommand{\qOneNorm}[1]{\left\| {#1} \right\|\idx{1}}
\newcommand{\qNorm}[1]{\left\| {#1} \right\|}
\newcommand{\qGivenNorm}{\qNorm{\boldsymbol{\cdot}}}

\newcommand{\qred}[1]{{\color{red}#1}}
\newcommand{\imesorange}{E77B29}
\newcommand{\imesgruen}{C8D317}
\newcommand{\imesblauHundert}{00509B}
\newcommand{\imesblauZwanzig}{CCDCEB}
\newcommand{\imesblauVierzig}{99B9D8}

\newcommand{\eg}{e.g.,\xspace}
\newcommand{\ie}{i.e.,\xspace}

\newcommand*{\MNIW}{\mathcal{MNIW}}
\newcommand*{\IW}{\mathcal{IW}}

\newcommand{\del}{\partial}
\newcommand{\bi}[1]{\boldsymbol{#1}}
\newcommand{\ur}[1]{\mathrm{#1}}
\newcommand{\cali}[1]{\mathcal{#1}}

\newcommand{\ubar}[1]{\underaccent{\bar}{#1}}
\newcommand{\ToDo}[1]{\todo[size=\tiny]{#1}}

\newcommand{\spans}[1]{\spanop\left( #1 \right)}

\newcommand{\ToDos}[1]{\todo[size=\tiny]{#1}}

\newcommand{\smFt}{\fontsize{4pt}{4pt}\selectfont}
\newcommand{\TTheta}{{\smFt $\mathrm{type}(\boldsymbol{\theta})$}}
\newcommand{\TX}{{\smFt $\mathrm{type}(\boldsymbol{x})$}}
\newcommand{\TY}{{\smFt $\mathrm{type}(\boldsymbol{y})$}}

\newcommand{\DOne}{{\smFt $\mathrm{dim}(1)$}}
\newcommand{\DTheta}{{\smFt $\mathrm{dim}(d_{\theta})$}}
\newcommand{\DX}{{\smFt $\mathrm{dim}(d_{x})$}}
\newcommand{\DY}{{\smFt $\mathrm{dim}(d_{y})$}}

\newcommand{\VTOne}{{\smFt $\mathrm{val}(\theta_{1})$}}
\newcommand{\VXOne}{{\smFt $\mathrm{val}(x^0_{1})$}}
\newcommand{\VYOne}{{\smFt $\mathrm{val}(y^0_{1})$}}
\newcommand{\VTdt}{{\smFt $\mathrm{val}(\theta_{d_{\theta}})$}}
\newcommand{\VXdx}{{\smFt $\mathrm{val}(x^T_{d_x})$}}
\newcommand{\VYdy}{{\smFt $\mathrm{val}(y^T_{d_y})$}}

\newcommand{\Zero}{{\smFt $0$}}
\newcommand{\One}{{\smFt $1$}}

\newcommand{\cobs}[1]{\textcolor[RGB]{170,0,0}{#1}}%
\newcommand{\cstat}[1]{\textcolor[RGB]{0,0,0}{#1}}%
\newcommand{\cpara}[1]{\textcolor[RGB]{0,170,0}{#1}}%

\newcommand{\lrw}{\texttt{lrw}\xspace}
\newcommand{\dw}{\texttt{dw}\xspace}
\newcommand{\clg}{\texttt{clg}\xspace}
\newcommand{\lv}{\texttt{lv}\xspace}

\newcommand{\subfig}[2]{\ref{#1}(#2)}   %

\newcommand{\unsure}[1]{\redcomment{[?~#1]}}   %
\newcommand{\fillin}[1]{\redcomment{[#1]}}     %

\begin{frontmatter}

\title{Conditional Diffusion Models for Amortized Bayesian System Identification\thanksref{footnoteinfo}}

\thanks[footnoteinfo]{This research was supported in part by the \textit{Kjell och Märta Beijer Foundation} and the \textit{Swedish Research Council (VR)} under the contracts 2021-04321 and 2025-04318, and the \textit{German Academic Scholarship Foundation (Studienstiftung des Deutschen Volkes)}.}

\author[LUH]{Mike Hänseroth}
\author[UU]{Niklas Wahlström}
\author[UU]{Thomas B. Schön}
\author[LUH,L3S]{Thomas Seel}
\author[LUH]{Jan-Hendrik Ewering}

\address[LUH]{Institute of Mechatronic Systems, Leibniz University Hannover, Germany (e-mail: mike.haenseroth@web.de, \{seel,ewering\}@imes.uni-hannover.de).}
\address[UU]{Dep. of Information Technology, Uppsala University, Sweden (e-mail: niklas.wahlstrom@it.uu.se, thomas.schon@uu.se).}
\address[L3S]{L3S Research Center, Leibniz University Hannover, Germany.}

\begin{abstract}

Generative diffusion models rapidly solve Bayesian inverse problems, such as image restoration and inpainting, with a single trained model and without retraining for each new observation or task.
Bayesian system identification, which typically involves inferring the hidden states and parameters of a dynamical system from measurements, poses a closely related inverse problem. 
Yet it is often addressed with computationally heavy Monte Carlo methods that must be rerun for every new measurement record. 
Moreover, these methods are restricted to a single inference task, and their cost grows with the state dimension and the record length. 
To alleviate these restrictions, we interpret Bayesian system identification in \acp{hmm} as a general Bayesian inverse problem, taking advantage of conditional diffusion models. 
Specifically, a transformer-based diffusion model is trained \emph{once} on short segments of simulated trajectories to approximate the joint posterior of states and parameters.
Through an attention mask induced by the Markov property and a relative time encoding, it can then be deployed on any new measurement record of custom length. 
At inference time, the same model can flexibly and rapidly be conditioned for joint state and parameter inference, state smoothing, parameter inference, or reconstructing missing measurements, without retraining.
We demonstrate these capabilities on four benchmark systems and examine how the length of the training segments affects state and parameter inference. 
On records five times longer than the training segments, the model maintains its state accuracy, whereas comparison methods either show a more than tenfold error increase or require ten times as much sampling time.

\end{abstract}


\end{frontmatter}
\raggedbottom
\acresetall

\section{Introduction}
\label{sec:intro}

Bayesian system identification~\citep{Chiuso.2019,Ljung.1999} often amounts to finding the posteriors of hidden system states, parameters, or both~\citep{Sarkka.2023,Wigren.2022,Schon.2011}.
Such problems arise, for instance, for a fleet of vehicles of the same type, which share one model but whose parameters differ from vehicle to vehicle because of manufacturing tolerances, wear, or varying environmental conditions, such as road friction.
For every vehicle and every change, new measurements can be recorded, and the inference tasks vary, from state smoothing to identifying the parameters or reconstructing missing measurements.
In this regard, it is desirable that a Bayesian system identification scheme possess three characteristics:
(i) rapid inference for any new measurement record,
(ii) flexible deployment to different inference tasks, and
(iii) generalization to measurement records of custom length. 

\begin{figure}[tb]
	\centering
	{\fontsize{6pt}{6pt}\selectfont
		\def\svgwidth{\columnwidth}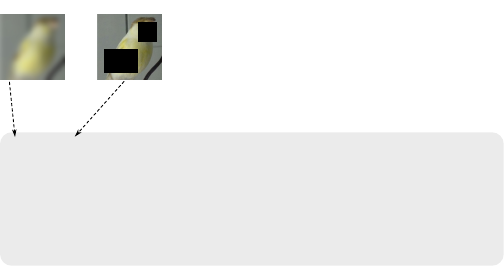}
	\caption{Image restoration and Bayesian system identification are both Bayesian inverse problems: unknown quantities are recovered from corrupted or incomplete observations through a known forward model. Amortized diffusion models can solve inference tasks in both domains without retraining. Inset: an amortized model pays its training cost once, after which each further task is cheap.}
	\label{fig:concept}
\end{figure}

Classical Bayesian inference schemes, such as \ac{pmmh}, a \ac{pmcmc} method~\citep{Andrieu.2010,Wigren.2022}, must be rerun for every new record, task, or pattern of missing measurements.
Neural \emph{amortized} inference methods, often associated with \ac{sbi}~\citep{Cranmer.2020}, avoid this: they pay the computational cost once in an offline training phase (see Fig.~\ref{fig:concept}, inset), after which no new record or task requires rerunning the entire inference pipeline.
However, most of them do not target Bayesian system identification, as they infer only parameters, and no hidden states, from observations of a fixed structure.
A rare exception is the ``Simformer''~\citep{Glockler.2024}, which samples arbitrary conditionals of a joint distribution, but does not distinguish hidden states from static parameters and has not been deployed on records longer than those seen in training.
Recent work addresses parts of these limitations, \eg \ac{sda}~\citep{Rozet.2023}, which targets state smoothing on records of arbitrary length with diffusion models, and its extension to parameter inference~\citep{Yamaoka.2026}.
However, to the best of our knowledge, no existing scheme amortizes the joint inference of states and parameters across inference tasks and record lengths, as detailed in Sec.~\ref{sec:Related_Work}, dedicated to related work.

To resolve these limitations, we take inspiration from image processing, in which generative diffusion models have shown great success in solving general Bayesian inverse problems, such as image restoration and inpainting~\citep{Luo.2023b,Luo.2023,Luo.2025b}, where a single trained model serves new observations and tasks without retraining~\citep{Zhao.2025,Luo.2024}.
Interestingly, these problems are closely related to the system identification tasks above, as illustrated in Fig.~\ref{fig:concept}: inpainting missing pixels, for instance, corresponds to reconstructing missing measurements.
Taking inspiration from this analogy, we pose Bayesian system identification in \acp{hmm} as a general Bayesian inverse problem. 
To solve it, we tailor the Simformer~\citep{Glockler.2024}, a diffusion transformer, to this setting.

Our first contribution is an amortized diffusion sampler to approximate the joint posterior of the states and parameters of \acp{hmm}; trained once on short segments of simulated trajectories, it returns posterior samples for new measurement records of custom length.
Second, the same trained model can be conditioned for several inference tasks without retraining, for instance, joint state and parameter inference, state smoothing, parameter inference, or reconstructing missing measurements.
We demonstrate these capabilities on four benchmark systems and examine how the length of the training segments affects state and parameter inference on records longer than the training segments.
Deployed on records up to five times the training length, the model keeps its state accuracy, whereas the state error of a dense-attention Simformer baseline can grow more than tenfold. 
Moreover, the proposed scheme samples an order of magnitude faster than \ac{pmmh}. 

\section{Problem Setting}
\label{sec:Problem_Setting}

To make the inference tasks of Sec.~\ref{sec:intro} precise, we first fix the considered class of systems.
Specifically, we assume that the data-generating system can be described by a known \ac{hmm} in state-space form,
\begin{align}
	\label{eq:HMMs}
	\boldsymbol{x}^{t} \mid \boldsymbol{x}^{t-1} &\sim p(\boldsymbol{x}^{t} \mid \boldsymbol{x}^{t-1}, \boldsymbol{\theta}), \\
	\boldsymbol{y}^{t} \mid \boldsymbol{x}^{t} &\sim p(\boldsymbol{y}^{t} \mid \boldsymbol{x}^{t},\boldsymbol{\theta}), \nonumber
\end{align}
with discrete time $t$, hidden states $\boldsymbol{x}^t$ (\eg positions, velocities, or population sizes), and noisy, possibly partial measurements $\boldsymbol{y}^t$.\footnote{The system may also be driven by known, non-random inputs $\boldsymbol{u}^t$. As is common in system identification, we keep them implicit in the densities, \ie the transition density at time $t$ depends on $\boldsymbol{u}^t$.}
A record covers $t=0,\dots,T$ and starts from an initial state $\boldsymbol{x}^0\sim p(\boldsymbol{x}^0\mid\boldsymbol{\theta})$.
The model is characterized by unknown, time-invariant parameters $\boldsymbol{\theta}$ (\eg masses or growth rates), for which we are given a prior $p(\boldsymbol{\theta})$.
By the Markov property, the joint distribution of all variables of a record factorizes as
\begin{align}
p(\boldsymbol{\theta},\boldsymbol{x}^{0:T},\boldsymbol{y}^{0:T})
  &= p(\boldsymbol{\theta})\,p(\boldsymbol{x}^0\mid\boldsymbol{\theta}) \label{eq:factorization} \\
  &\quad\cdot\textstyle\prod_{t=1}^{T} p(\boldsymbol{x}^{t}\mid\boldsymbol{x}^{t-1},\boldsymbol{\theta})\prod_{t=0}^{T} p(\boldsymbol{y}^t\mid\boldsymbol{x}^t,\boldsymbol{\theta}). \nonumber
\end{align}

Given this system class, different inference tasks condition on different parts of a measurement record.
With all variables collected in $\boldsymbol{v}=(\boldsymbol{\theta},\boldsymbol{x}^{0:T},\boldsymbol{y}^{0:T})$, a task specifies which entries are observed, $\boldsymbol{v}_{\mathrm{o}}$, and asks for the posterior of the unknown ones, $p(\boldsymbol{v}_{\mathrm{u}}\mid\boldsymbol{v}_{\mathrm{o}})\propto p(\boldsymbol{v}_{\mathrm{u}},\boldsymbol{v}_{\mathrm{o}})$, so that every task is a conditional of the same joint~\eqref{eq:factorization}. 
Relevant inference problems are, for instance,
\begin{equation}
\label{eq:task_posterior}
\begin{aligned}
&\text{joint inference:} && p(\boldsymbol{\theta},\boldsymbol{x}^{0:T}\mid\boldsymbol{y}^{0:T}),\\
&\text{state smoothing:} && p(\boldsymbol{x}^{0:T}\mid\boldsymbol{\theta},\boldsymbol{y}^{0:T}),\\
&\text{parameter inference:} && p(\boldsymbol{\theta}\mid\boldsymbol{x}^{0:T},\boldsymbol{y}^{0:T}),\\
&\text{reconstruction:} && p(\boldsymbol{\theta},\boldsymbol{x}^{0:T},\boldsymbol{y}_{\mathrm{m}}\mid\boldsymbol{y}_{\mathrm{o}}),
\end{aligned}
\end{equation}
where the reconstruction task infers missing outputs $\boldsymbol{y}_{\mathrm{m}}$ together with states and parameters from irregularly observed measurements, $\boldsymbol{y}_{\mathrm{o}}$. 

The objective in this paper is to find an inference scheme that enables solving various inference tasks, such as \eqref{eq:task_posterior}, without rerunning the entire inference pipeline. 
In particular, this objective is characterized by the following three challenges:
\begin{itemize}
	\item[(i)] \emph{Rapid, amortized Bayesian system identification:} we require a computationally efficient deployment, such that inference can be performed rapidly for new records;
	\item[(ii)] \emph{Flexible inference tasks:} the scheme should be flexibly applicable to different inference objectives; 
	\item[(iii)] \emph{Variable deployment length $T$:} the method should generalize to custom record lengths $T$, as these may vary in a given application.
\end{itemize}

\section{Background}
\label{sec:Background}
As outlined in Sec.~\ref{sec:intro}, we treat all tasks in~\eqref{eq:task_posterior} as a Bayesian inverse problem and approximate their posteriors with one conditional diffusion model, designed to tackle the three challenges above.

\emph{First}, to enable rapid inference, we follow the \ac{sbi} paradigm and shift the computational effort to an offline training phase on simulations of~\eqref{eq:HMMs}.
Loosely speaking, given $N$ evaluations of the model~\eqref{eq:HMMs}, \ac{pmcmc} spends all of them on a single inference task and needs the same number again for any other measurement record. In contrast, we propose to leverage the $N$ evaluations upfront in offline training, making the contained information reusable for any new measurement record and inference task (Fig.~\ref{fig:concept}, inset).
\emph{Second}, to serve different inference tasks, we build on random conditioning masks~\citep{Glockler.2024}, which encode a broad space of possible $\boldsymbol{v}_\mathrm{o}$ and $\boldsymbol{v}_\mathrm{u}$ during training.
\emph{Third}, since the record length at deployment is unknown, we train on short windows of trajectories and deploy on full records, enabled by a Markov attention mask and a relative time encoding.

To provide suitable context for these adaptations, Secs.~\ref{subsec:Conditional_Diffusion_Samplers} and~\ref{subsec:The_Simformer_Architecture} review conditional diffusion samplers and the Simformer.

\subsection{Conditional Diffusion Samplers}
\label{subsec:Conditional_Diffusion_Samplers}
Diffusion models sample from a potentially complex target distribution by drawing from a simple distribution and reversing a gradual noising process~\citep{Ho.2020,Song.2021}.
For a variable $\boldsymbol{z}$ with conditioning information $\boldsymbol{c}$ (in our setting $\boldsymbol{z}=\boldsymbol{v}_{\mathrm{u}}$ and $\boldsymbol{c}=\boldsymbol{v}_{\mathrm{o}}$), the forward process perturbs only $\boldsymbol{z}$, while $\boldsymbol{c}$ is kept fixed.
We use the variance-exploding \ac{sde}~\citep{Song.2021}, whose perturbation kernel is Gaussian,
\begin{equation}
\label{eq:pert_kernel}
p_a(\boldsymbol{z}_a\mid\boldsymbol{z}_0)=\mathcal{N}\big(\boldsymbol{z}_0,\sigma^2(a)\boldsymbol{I}\big),
\end{equation}
\ie $\boldsymbol{z}_a=\boldsymbol{z}_0+\sigma(a)\boldsymbol{\epsilon}$ with $\boldsymbol{\epsilon}\sim\mathcal{N}(\boldsymbol{0},\boldsymbol{I})$.
The noise scale $\sigma(a)$ increases over the diffusion time $a\in[0,A]$ (written $a$ to distinguish it from the sequence time $t$) from $\sigma(0)\approx0$ to a $\sigma(A)$ large enough such that $\boldsymbol{z}_A$ is approximately Gaussian.
\citet{Anderson.1982} showed that the reverse-time process is the \ac{sde}
\begin{equation}
\label{eq:reverse_sde}
\mathrm{d}\boldsymbol{z}_a=-g^2(a)\nabla_{\boldsymbol{z}_a}\log p_a(\boldsymbol{z}_a\mid\boldsymbol{c})\,\mathrm{d}a+g(a)\,\mathrm{d}\tilde{\boldsymbol{w}},
\end{equation}
with $g^2(a)=\mathrm{d}\sigma^2(a)/\mathrm{d}a$ and a reverse-time Wiener process $\tilde{\boldsymbol{w}}$.
Its drift contains the score $\nabla_{\boldsymbol{z}_a}\log p_a(\boldsymbol{z}_a\mid\boldsymbol{c})$, which is intractable.
The score of the kernel~\eqref{eq:pert_kernel}, however, is known, $\nabla_{\boldsymbol{z}_a}\log p_a(\boldsymbol{z}_a\mid\boldsymbol{z}_0)=-\boldsymbol{\epsilon}/\sigma(a)$, so a network $s_\psi(\boldsymbol{z}_a,\boldsymbol{c},a)$ can be trained by denoising score matching~\citep{Vincent.2011,Song.2021} with the objective
\begin{equation}
\label{eq:loss_fn2}
\mathcal{L}(\psi)=\mathbb{E}_{a,\boldsymbol{z}_0,\boldsymbol{c},\boldsymbol{\epsilon}}\big[\lambda(a)\,\|s_\psi(\boldsymbol{z}_a,\boldsymbol{c},a)+\boldsymbol{\epsilon}/\sigma(a)\|_2^2\big],
\end{equation}
where $\lambda(a)>0$ is a weighting function and $a\sim\mathcal{U}(0,A)$.
For a sufficiently expressive network, the minimizer of~\eqref{eq:loss_fn2} is the intractable conditional score $\nabla_{\boldsymbol{z}_a}\log p_a(\boldsymbol{z}_a\mid\boldsymbol{c})$~\citep{Vincent.2011}.
Samples are then drawn by integrating~\eqref{eq:reverse_sde} from $a=A$ to $0$ with the learned score.

\subsection{The Simformer Architecture}
\label{subsec:The_Simformer_Architecture}

The Simformer~\citep{Glockler.2024}, which we revisit in the following, provides the methodological basis of our model.
Like other \ac{sbi} methods, it is amortized, \ie trained once on simulated data, but unlike most of them, which target the posterior or the likelihood, it approximates the joint distribution $p(\boldsymbol{\theta},\boldsymbol{y})$ of parameters and data, and thereby all of its conditionals, with a masked transformer that serves as the score network of a conditional diffusion sampler as in Sec.~\ref{subsec:Conditional_Diffusion_Samplers}.
This architecture is convenient for system identification because every variable, \eg each entry of each time step, is a separate token processed by the same weights, so the network is not tied to a fixed input dimension and can, in principle, process records of different lengths.

Its first key feature is that a single trained model samples arbitrary conditionals of the joint distribution, \eg the posterior of some parameters given the remaining ones and the data.
To this end, a conditioning mask $\boldsymbol{m}_C$ over all tokens marks conditioned variables with $m_{C,i}=1$ and variables to be inferred with $m_{C,i}=0$.
During training, \citet{Glockler.2024} draw the conditioning mask uniformly at random from the joint, posterior, and likelihood masks and two random Bernoulli masks.
Training across conditioning patterns suits system identification, where the tasks in~\eqref{eq:task_posterior} differ only in which $\boldsymbol{v}_\mathrm{o}$ variables are observed, and which other variables $\boldsymbol{v}_\mathrm{u}$ are to be inferred.

A second key feature is that attention can be restricted to encode known dependencies: masking the attention operation prevents selected tokens from attending to specific others, which limits the receptive field of the transformer.
This suits dynamical systems, whose factorization~\eqref{eq:factorization} prescribes such dependencies, as \citet{Glockler.2024} demonstrate on a short \ac{hmm} measurement record.
For further details on the architecture, we refer to \citet{Glockler.2024}.

For system identification, however, the Simformer has two limitations.
First, it distinguishes only parameters and data, so neither hidden states nor the known inputs that drive most systems in system identification have a role of their own.
In principle, the Simformer could distinguish hidden states from measurements through diffusion guidance~\citep{Glockler.2024}, with the measurement model entering only at inference time.
It is not, however, tailored to \acp{hmm} or Bayesian system identification, so neither joint state and parameter inference nor the structure of~\eqref{eq:factorization}, in which static parameters couple all time steps, is encoded or studied.
Second, its Fourier embedding of the absolute time permits evaluation at custom time points, but records longer than those seen in training then contain times outside the training distribution.
Deployment on such records, and thus the accuracy of inference beyond the training horizon, is not studied.
The subsequent section addresses these limitations and presents a flexible scheme for amortized Bayesian system identification.

\section{Methods}
\label{sec:Methods}
\begin{figure}[tb]
	\centering
	{\fontsize{5pt}{6pt}\selectfont
		\def\svgwidth{\columnwidth}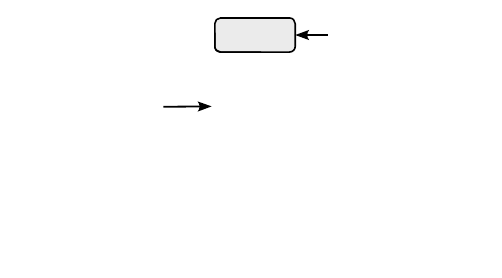}
	\caption{The Markov mask encodes the dependencies of the \ac{hmm} in the attention. All shaded entries are attended to, and the shades mark how the two tokens are related (legend). Since this pattern repeats from step to step, it applies unchanged to records of custom length. Adapted from \citet{Glockler.2024}.}
	\label{fig:architecture}
\end{figure}

In this section, we adapt the Simformer to Bayesian system identification (Fig.~\ref{fig:architecture}).
Specifically, we address its two limitations in turn: First, we introduce separate token types and encode the dependency structure of~\eqref{eq:factorization} in the attention mask (Sec.~\ref{subsec:Markov_Mask}).
Second, we encode time such that the model can be deployed on records of custom length (Sec.~\ref{subsec:RoPE_and_GFE}). 
Since these adaptations decouple the model from the record length, it can be trained on short windows of the simulated trajectories.
Section~\ref{subsec:Width_Grid} introduces this windowed training and the hypotheses on its effect that we test in Sec.~\ref{sec:Results}.

\subsection{Markov Mask}
\label{subsec:Markov_Mask}

Unlike the Simformer, which tokenizes parameters and data only, we tokenize parameters, hidden states, measurements, and inputs as separate token types (see Fig.~\ref{fig:architecture}), so that the dependencies of~\eqref{eq:factorization} can be encoded in the attention mask.
Under this factorization, each state is linked directly only to its neighbouring states, its own measurement and input, and the parameters $\boldsymbol{\theta}$, whereas $\boldsymbol{\theta}$ couples the entire record.
To reflect this in the model, we restrict attention to exactly these links (Fig.~\ref{fig:architecture}) and call the resulting attention mask, which specializes the Simformer's dependency masks to~\eqref{eq:factorization}, the ``Markov mask''.
The mask is symmetric and thus encodes the undirected neighbourhood of each variable in the graph of Fig.~\ref{fig:windowedTrainingSchematic} (top), as suits smoothing, where each state depends on both its predecessor and its successor. %
Within each layer, the state at time $t$ attends to the previous and next state.
Over $L$ layers, the receptive field of each state thus covers $L$ neighbouring states in both the forward and backward temporal directions. 
Information can, however, also pass between distant states through the parameter tokens, so the Markov mask does not impose a hard constraint. 
Besides introducing an inductive bias consistent with the dynamical system, the Markov mask lets attention be evaluated with a banded kernel, whose cost grows linearly with the record length $T$ rather than quadratically, as for the dense attention of the Simformer~\citep{Glockler.2024}, in which every token attends to every other.

\subsection{Temporal Encoding}
\label{subsec:RoPE_and_GFE}

To deploy the model on records of custom length, time must be encoded independently of the record length.
\citet{Glockler.2024} propose to embed the time of a token through a \ac{gfe} of its absolute value, \ie sinusoids at fixed random frequencies.
However, with this encoding, records longer than the simulated training trajectories contain absolute times whose embeddings were not present in training.
Therefore, we encode the time $t$ of each token with \ac{rope}~\citep{Su.2024} inside the attention mechanism, where time enters only through temporal differences between tokens.
Since the Markov mask limits these differences to at most one time step per layer, every difference that occurs at deployment was seen in training, such that the time encoding is in-distribution, regardless of the absolute time.
Parameter tokens carry no time, and therefore parameter-related scores are not rotated.

\subsection{Windowed Training}
\label{subsec:Width_Grid}
\begin{figure}[tb]
	\centering
	{\fontsize{7pt}{7pt}\selectfont
		\def\svgwidth{\columnwidth}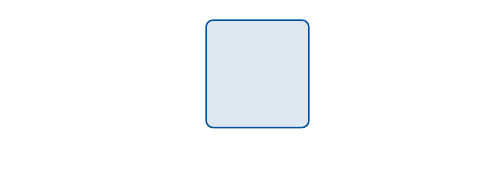}
	\caption{The model is trained on short windows of simulated trajectories, of fixed or randomly drawn width, and applied unchanged to a longer measurement record: each state depends only on its neighbourhood, whereas the parameters $\boldsymbol{\theta}$ couple all time steps. Black edges lie inside the window; graph adapted from \citet{Wigren.2022}.}
	\label{fig:windowedTrainingSchematic}
\end{figure}

Since the architecture allows records of flexible length at inference time, we can train on short segments and deploy on full records. 
We therefore draw windows of $|W|$ consecutive time steps from the simulated trajectories, with the width either fixed or drawn anew for each batch, and apply the trained model to the full record at inference time (see Fig.~\ref{fig:windowedTrainingSchematic}).
A state whose receptive field lies inside the window receives the same neighbouring states and measurements as in the full record, and since the time encoding depends only on time differences, the position of the window within the record does not matter.
For the parameters, which couple the entire record, the window is the only evidence available during training, so the width sets how much evidence about $\boldsymbol{\theta}$ the model learns to aggregate.

This reasoning leads to two hypotheses, which we test in Sec.~\ref{sec:Results}.
(H1)~Because the influence of a measurement on a state decays with the time between them---a forgetting property that underlies various classical estimation schemes, such as fixed-lag smoothing~\citep{Sarkka.2023}---short windows should suffice to infer the hidden states, whereas the accuracy of the parameters should keep improving with the width.
(H2)~Under the Markov mask, each state attends only to its neighbours, its own measurement and input, and the parameters $\boldsymbol{\theta}$.
We therefore expect the state accuracy to carry over to records longer than the training trajectories.
For the parameters, longer records contract the exact posterior, \ie make it narrower.
We expect the parameter posterior of our model not to contract accordingly, because it learns to combine evidence about $\boldsymbol{\theta}$ only over windows as long as those seen in training.

\clearpage
\section{Empirical Results}
\label{sec:Results}

Having introduced the adapted Simformer, we now evaluate it against the three characteristics of Sec.~\ref{sec:intro} and test the hypotheses of Sec.~\ref{subsec:Width_Grid}.
After describing the study design, we demonstrate that one trained model serves several inference tasks (Sec.~\ref{subsec:Capabilities}).
We then deploy it on records longer than the training trajectories and attribute the results to the components of Sec.~\ref{sec:Methods} (Sec.~\ref{subsec:Deployment}).
Finally, we examine how the width of the training windows affects state and parameter inference (Sec.~\ref{subsec:Width}).

\subsection{Study Design}
\label{subsec:Study_Design}
To cover linear, nonlinear, and input-driven dynamics, we consider four benchmark systems, for which reference posteriors are available and in which all states are measured with Gaussian noise of variance $\sigma_y^2$:
\begin{itemize}
	\item \emph{Linear random walk} (\lrw)~\citep{Glockler.2025}: $\boldsymbol{x}^t=0.9\,\boldsymbol{x}^{t-1}+\boldsymbol{\theta}+\boldsymbol{\eta}^t$ with $d_x=d_\theta=2$ and $\sigma_y=0.5$; closed-form reference;
	\item \emph{Double well} (\dw), adapted from~\citet{Glockler.2025}: $\mathrm{d}x=(\theta_1x-\theta_2x^3)\,\mathrm{d}t+\mathrm{d}w$ with $\boldsymbol{\theta}\sim\mathcal{U}([0.5,3]^2)$, discretized with step $0.1$, for each of the $d_x=2$ components with shared $\boldsymbol{\theta}$ and $\sigma_y=0.5$; grid-filter reference;
	\item \emph{Controlled linear-Gaussian} (\clg): $x^t=a\,x^{t-1}+b\,u^t+\eta^t$ with known inputs $u^t$ and $\boldsymbol{\theta}=(a,b)$, where $b$ is identifiable only through the inputs, and $\sigma_y=0.5$; exact grid reference;
	\item \emph{Lotka--Volterra} (\lv)~\citep{Glockler.2024}: deterministic predator--prey dynamics with $d_x=2$ and $\boldsymbol{\theta}=(\alpha,\beta,\gamma,\delta)$, and $\sigma_y=0.1$; \ac{pmmh} reference~\citep{Andrieu.2010}.
\end{itemize}

All models use the architecture of Sec.~\ref{sec:Methods} with $L=8$ layers, $4$ attention heads, tokens of width $96$, and a feed-forward width of $384$.
They are trained with Adam~\citep{kingma2017} at learning rate $10^{-3}$ on simulated trajectories of length $T_{\mathrm{train}}=50$, with early stopping on a $20\,\%$ validation split (patience of $200$ epochs).
Unless stated otherwise, all configurations see the same budget of $10^5$ simulated time steps and use the mixed width scheme of Sec.~\ref{subsec:Width}.
It draws the width $|W|\sim\mathcal{U}\{17,\dots,50\}$ anew for each batch, so that no window truncates the receptive field of $2L+1=17$ time steps.
The batch size scales inversely with the width, the largest one for the mixed scheme, so that all configurations take the same number of optimizer steps per epoch, and the loss weights of parameters and states are balanced per training sample.
During training, the conditioning mask selects the observed variables $\boldsymbol{v}_{\mathrm{o}}$, and drawing it at random trains a single model for all tasks in~\eqref{eq:task_posterior}.
We compare a random mask, which marks each variable independently as observed with probability $0.2$, with a mixture of five masks, one per task in~\eqref{eq:task_posterior} plus the random mask, drawn uniformly for each training sample.
Since both yield about the same accuracy for joint inference (see Fig.~\ref{fig:iii_sel_factors}C), we use the more general random mask throughout.
\ac{pmmh} is run separately on every record, with four chains of $15\,000$ iterations, $5\,000$ of them burn-in, on \lrw, \clg, and \dw, and with four chains of $40\,000$ iterations, $10\,000$ of them burn-in, on \lv; it provides the reference for \lv and serves as a non-amortized comparison method on the other systems, scored like our model on $500$ draws thinned from its chains.
Both run on GPUs; all sampling times are measured on one NVIDIA L40S.

\begin{figure}[tb]
	\centering
	{\fontsize{7pt}{7pt}\selectfont
		\def\svgwidth{\columnwidth}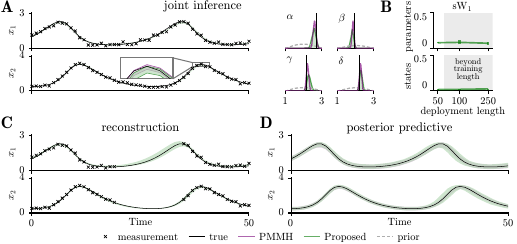}
	\caption{One trained model solves, without retraining, joint inference in good agreement with \ac{pmmh} (A), and reconstruction of missing measurements (C). In (D), we give forward simulations with parameters from the joint posteriors, both for \ac{pmmh} and the proposed method. Its parameter error does not grow and its state error grows by about half on records up to five times the training length (B). Bands: $90\,\%$ credible intervals, in (D) $95\,\%$ predictive intervals, from $2000$ posterior samples per record. Error bars in (B): $95\,\%$ $t$-intervals over five repetitions.}
	\label{fig:iii_lv_showcase}
\end{figure}

Unless stated otherwise, we target the joint $p(\boldsymbol{\theta},\boldsymbol{x}^{0:T}\mid\boldsymbol{y}^{0:T})$ on ten fixed test records of length $T=50$ per system, simulated with parameters drawn from the prior and shared by all configurations.
Each configuration is trained five times, each with a fresh training set and initialization.
We draw $500$ posterior samples per record by integrating~\eqref{eq:reverse_sde} in $100$ Euler--Maruyama steps and compare them with samples from the reference posterior of each system, separately for the parameters and the states.
To quantify accuracy, we use the sliced Wasserstein distance $\mathrm{sW}_1$~\citep{Bonneel.2015}, which averages the one-dimensional Wasserstein distance over $128$ random projections and is zero for identical distributions.
Results are averaged over the ten records and reported as means with $95\,\%$ $t$-intervals over the trained models. %

\subsection{One Model, Several Tasks}
\label{subsec:Capabilities}
To demonstrate characteristic (ii), the flexible deployment to different inference tasks, we consider the proposed approach on \lv (Fig.~\ref{fig:iii_lv_showcase}).
Of the tasks in~\eqref{eq:task_posterior}, we show joint inference, conditioned on all measurements of a record (A), and reconstruction, with $15$ consecutive measurements removed from the same record (C).
The same model solves both without retraining.
For joint inference, the state posterior (A, left) agrees with the \ac{pmmh} reference, and the parameter posterior (A, right) agrees in location, but is wider, and forward simulations with parameters drawn from this posterior reproduce the true trajectory (D).
For reconstruction, the model fills the gap with a posterior band that contains the true trajectory (C).
State smoothing and parameter inference tasks can be readily solved in a similar fashion, by additionally conditioning on the parameters or the states.
This illustrates the practical gain of amortizing over tasks: switching between them only changes the conditioning mask (Sec.~\ref{subsec:The_Simformer_Architecture}), whereas \ac{pmmh} needs to run the entire inference pipeline for every record and every pattern of missing measurements.
On records of up to $T=250$, the parameter error does not grow and the state error grows by about half (B), which Sec.~\ref{subsec:Deployment} examines next.

\begin{figure}[tb]
	\centering
	{\fontsize{7pt}{7pt}\selectfont
		\def\svgwidth{\columnwidth}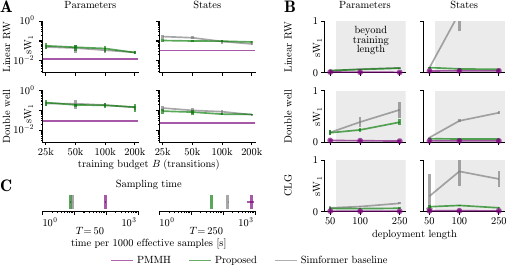}
	\caption{Beyond the training length $T_{\mathrm{train}}=50$, our model keeps its state accuracy, whereas the Simformer baseline degrades, and only \ac{pmmh} gains parameter accuracy on longer records (B, linear scale). In exchange, our model needs about a twelfth to a seventeenth of the sampling time of \ac{pmmh} (C). Error bars: $95\,\%$ $t$-intervals over five repetitions.}
	\label{fig:iii_toy_study}
\end{figure}
\subsection{Deployment on Longer Records}
\label{subsec:Deployment}
To test (H2), we apply the models trained with the mixed scheme on $T_{\mathrm{train}}=50$ to records of length $T\in\{50,100,250\}$ (Fig.~\ref{fig:iii_toy_study}).
We compare them with \ac{pmmh} and with a Simformer baseline~\citep{Glockler.2024}, which uses dense attention and \ac{gfe}, is trained without windowing, and otherwise shares our tokenization, conditioning masks, and training budget.
At the training length, more training data does not close the gap to \ac{pmmh} (A): an eightfold training budget lowers the error of our model by $16$--$55\,\%$, but even at the largest budget, its parameter error remains about two to five times that of \ac{pmmh}.

Beyond the training length (B), the results support (H2).
On the states, our model keeps its accuracy at all deployment lengths, whereas the state error of the baseline grows more than tenfold on \lrw.
On the parameters, our model is more accurate than the baseline on \dw and \clg and on par on \lrw.
Unlike \ac{pmmh}, however, it does not turn the additional evidence of longer records into a narrower posterior, since it learned to aggregate this evidence only within its training windows (Sec.~\ref{subsec:Width_Grid}).
Its parameter error therefore roughly doubles from $T=50$ to $T=250$ on \lrw and \dw.

While the proposed method is thus an approximation that is less accurate than the asymptotically exact \ac{pmmh}, it samples more than ten times faster, needing $7$\,s instead of $94$\,s at $T=50$ and $43$\,s instead of $760$\,s at $T=250$ per $1000$ effective samples on \lrw (C).
It is also about three times faster than the baseline at $T=250$, as expected from the quadratic cost of dense attention (Sec.~\ref{subsec:Markov_Mask}).
Effective samples are the independent posterior samples for the diffusion models and, for \ac{pmmh}, the autocorrelation-corrected effective sample size of its chains.
For the states, our model thus meets characteristics (i) and (iii) of Sec.~\ref{sec:intro}, at the price of parameter posteriors that do not contract beyond the training windows.

\begin{figure}[tb]
	\centering
	{\fontsize{7pt}{7pt}\selectfont
		\def\svgwidth{\columnwidth}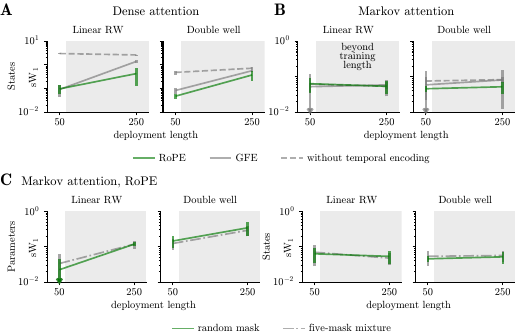}
	\caption{Under dense attention, the state error grows beyond the training length for every time encoding, least with \ac{rope} (A), whereas under the Markov mask, it stays constant (B). With the Markov mask and \ac{rope}, the five-mask mixture performs like the random mask used throughout (C). Models trained without windowing. Error bars: $95\,\%$ $t$-intervals over three repetitions.}
	\label{fig:iii_sel_factors}
\end{figure}

To explain this behaviour, and since the baseline differs from our model in the attention mask, the time encoding, and the windowing at once, we isolate the first two by training all combinations of attention mask (Markov, dense) and time encoding (\ac{rope}, \ac{gfe}, none) without windowing, like the baseline, on \lrw and \dw, with three repetitions each. %
As shown in Fig.~\ref{fig:iii_sel_factors}, the Markov mask, not the time encoding, carries the state accuracy beyond the training length. %
Under dense attention, the state error grows from $T=50$ to $T=250$ with either encoding, least with \ac{rope}, and is high from the start without one (A), whereas under the Markov mask, it stays nearly constant for all encodings (B).
In this ablation, the Markov mask thus accounts for the state part of (H2), and it reflects the structure of \acp{hmm}.
Under the Markov mask, each state attends only to the variables it is linked to in~\eqref{eq:factorization}, so a longer record adds tokens without changing the direct neighbourhood of any state.
The mask itself also encodes which states are neighbours, which may explain why even no time encoding suffices.
Under dense attention, in contrast, lengthening the record changes the input of every state, and a relative encoding only limits the resulting growth, whereas an absolute encoding has to extrapolate to times never seen in training.
Unlike the attention mask, the conditioning masks show no difference for joint inference: the five-mask mixture and the random mask yield overlapping intervals at both deployment lengths (C).
Whether the mixture helps for the other tasks in~\eqref{eq:task_posterior} remains to be tested, although the random mask already suffices for the reconstruction in Sec.~\ref{subsec:Capabilities}.
For both conditioning masks, the parameter error grows with $T$, as (H2) predicts, since the reference posterior contracts while that of the model does not.

\subsection{Ablation of the Training Window Width}
\label{subsec:Width}
\begin{figure}[tb]
	\centering
	{\fontsize{7pt}{7pt}\selectfont
		\def\svgwidth{\columnwidth}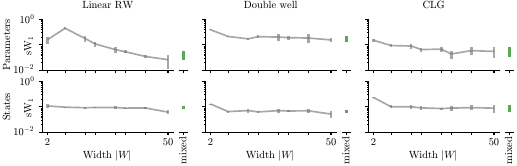}
	\caption{Wider training windows mainly help the parameters, which couple the entire record, whereas the states are accurate from short windows on. Mixed: $|W|\sim\mathcal{U}\{17,\dots,50\}$. Error bars: $95\,\%$ $t$-intervals over five repetitions.}
	\label{fig:iii_width}
\end{figure}

Finally, to test (H1) and to justify the mixed scheme used so far, we train models with fixed widths $|W|\in\{2,9,17,21,29,33,41,50\}$, where $|W|=50$ amounts to training without windowing, and with the mixed scheme, all with the same budget, and evaluate them at $T=50$ (Fig.~\ref{fig:iii_width}).
In line with (H1), the states need only short windows: from $|W|=9$ on, their accuracy is flat on \lrw, \clg, and \dw, and only single transitions ($|W|=2$) fall short.
The parameters, in contrast, gain from longer windows, up to $|W|\approx17$--$21$ on \dw and \clg, and up to $|W|=41$ on \lrw.
We attribute this asymmetry to the structure of \acp{hmm}.
A state is informed mainly by nearby measurements, whereas the evidence about $\boldsymbol{\theta}$ is spread over the whole record.
Given the states, the likelihood factorizes over the time steps~\eqref{eq:factorization}, but at inference time, only the noisy measurements are known, and the states are perturbed by the diffusion process together with the parameters, so this factorization no longer holds~\citep{Glockler.2025}.
The model therefore has to learn how to aggregate the evidence about $\boldsymbol{\theta}$ over the record, and it learns this only within its training windows.
In practice, the mixed scheme is about as accurate as the most accurate fixed widths and thus requires no choice of the width.

\section{Related Work}
\label{sec:Related_Work}
To position our scheme, we review the state of the art with respect to the three characteristics of Sec.~\ref{sec:intro}.
First, classical Bayesian system identification schemes, such as \ac{pmcmc}, target the joint posterior of states and parameters~\citep{Andrieu.2010,Wigren.2022,Volkmann.2025}, and Bayesian filters and smoothers infer the states for given parameters~\citep{Sarkka.2023}.
For nonlinear or non-Gaussian systems, however, they rely on computationally heavy Monte Carlo sampling, whose cost grows with the state dimension, and particle methods degenerate on long records~\citep{Andrieu.2010,Wigren.2022,Wills.2023}.
Moreover, they are not amortized, so every new record, task, or pattern of missing measurements requires a new execution of the \textit{entire} inference pipeline.
In this regard, \ac{sda}~\citep{Rozet.2023} provides amortized state inference: it trains a diffusion model once and, by decomposing the trajectory score into short segments, applies it to trajectories of arbitrary length, but it cannot perform parameter inference.
Building on it, \citet{Yamaoka.2026} augment the states with the parameters, in the spirit of self-organizing state-space models, and thereby amortize the joint inference of states and parameters.
However, their approach considers no other inference task. 
In contrast, our transformer represents the parameters by tokens of their own that attend to the entire record, and encodes the Markov structure of \acp{hmm} in its attention mask. 
In system identification, in-context learning~\citep{Forgione.2023} also trains a single transformer on simulations of a class of systems, but it predicts outputs rather than inferring the posterior of states and parameters of a given model.

Second, beyond classical system identification approaches, an \emph{amortized} approach to Bayesian inference is prevalent in \ac{sbi}, which considers models that can be simulated but whose likelihood is intractable~\citep{Cranmer.2020,ZammitMangion.2025}.
Specifically, recent \ac{sbi} methods train neural networks on simulations to approximate the likelihood, a likelihood ratio, or the posterior~\citep{Cranmer.2020,Lueckmann.2021}. 
Posterior estimators then return the posterior for any new observation without rerunning the simulator~\citep{Papamakarios.2016}, and diffusion models have recently shown strong results in this role~\citep{Sharrock.2024,Arruda.2025}.
However, \ac{sbi} treats the simulator as a ``black box''~\citep{Cranmer.2020} and does not infer hidden system states. 
Moreover, most approaches infer parameters from observations of a fixed structure, so missing measurements or a different record length require retraining~\citep{Glockler.2024}. 
Compositional \ac{sbi} for time series~\citep{Glockler.2025} trains on single transitions and composes the learned scores to infer the parameters of records of arbitrary length, but it does not infer the hidden states.
In contrast, the Simformer~\citep{Glockler.2024}, on which we build (Sec.~\ref{subsec:The_Simformer_Architecture}), samples arbitrary conditionals of the joint distribution of parameters and simulated data and thus serves different inference tasks, including the reconstruction of missing data, without retraining.
As detailed in Sec.~\ref{subsec:The_Simformer_Architecture}, however, it represents neither hidden states nor known inputs natively, its dense attention scales quadratically with the record length, and its deployment on records of other lengths is not studied.
Thus, none of these schemes combines all three characteristics: (i) rapid, amortized Bayesian system identification, (ii) flexible deployment to different inference tasks, and (iii) generalization to measurement records of custom length.

\section{Conclusions}
\label{sec:Conclusions}

In this paper, we exploit amortized conditional diffusion models, typically used in complex image processing tasks, for amortized and flexible Bayesian system identification. 
This approach is motivated by the idea that Bayesian system identification in \acp{hmm} can be interpreted as a general Bayesian inverse problem. 

Methodologically, we tailor a recently proposed simulation-based inference method, the Simformer for use in \acp{hmm} by a Markov attention mask and a relative time encoding. 
This enables the proposed sampler to perform different inference tasks, including joint state and parameter estimation and sequence reconstruction, on any new observation record of custom length. 
In contrast to \ac{pmmh}, it uses a number of model evaluations upfront during offline training, and \emph{amortizes} at inference time by rapid posterior sampling. %

The results show that the proposed approach keeps its state accuracy on records up to five times longer than the training trajectories, whereas the state error of the baseline Simformer can grow more than tenfold. 
While the proposed method is an approximation that provides reasonable accuracy compared to asymptotically exact \ac{pmmh}, it provides posterior samples at least one order of magnitude faster. 
This effect grows with more observation records and longer sequence lengths, which we attribute to the parallelized, non-Markov-chain sampling procedure. 
Future work should focus on aggregating parameter evidence beyond the training windows, and analyzing inference in high-dimensional systems.

To conclude, we present a new paradigm for Bayesian system identification with diffusion models that complements classical \ac{pmcmc} approaches for problem settings that admit amortization over multiple observation records. 
Moreover, diffusion-based identification approaches, being particle-free, have the potential to overcome limitations of \ac{pmcmc}, such as path degeneracy and poor mixing.

\bibliography{refs}

\end{document}